\documentclass{article}

\usepackage{journal2e_v1}
\usepackage[a4paper]{geometry}

\usepackage{amssymb}
\usepackage{booktabs}
\usepackage{lmodern}
\usepackage{mathtools}
\usepackage{nth}
\usepackage{graphicx}
\usepackage{subfigure}
\usepackage{mathrsfs}

\makeindex

\def\A{{\bf A}}

\def\B{{\bf B}}
\def\b{{\bf b}}

\def\c{{\bf c}}
\def\D{{\bf D}}

\def\E{{\bf E}}

\def\G{{\bf G}}
\def\I{{\bf I}}

\def\L{{\bf L}}

\def\R{{\bf R}}
\def\X{{\bf X}}
\def\Y{{\bf Y}}

\def\S{{\bf S}}
\def\T{{\mathcal T}}
\def\x{{\bf x}}

\def\Z{{\bf Z}}
\def\M{{\bf M}}

\def\N{{\bf N}}

\def\U{{\bf U}}
\def\u{{\bf u}}
\def\V{{\bf V}}

\def\0{{\bf 0}}
\def\1{{\bf 1}}
\def\wrt{{w.r.t.\ }}

\makeatletter
\newcommand\incircbin
{%
  \mathpalette\@incircbin
}

\newcommand\@incircbin[2]
{%
  \mathbin%
  {%
    \ooalign{\hidewidth$#1#2$\hidewidth\crcr$#1\bigcirc$}%
  }%
}

\makeatother

\DeclarePairedDelimiter{\floor}{\lfloor}{\rfloor}

\def\nnz#1{\|#1\|_0}

\def\norm#1{\|#1\|}

\def\AM{{\mathcal A}}
\def\BM{{\mathcal B}}
\def\EM{{\mathcal E}}

\def\RB{{\mathbb R}}
\def\FB{{\mathbb F}}

\def\rank{\operatorname{rank}}
\def\fold{\operatorname{fold}}

\def\vect{\operatorname{vec}}

\begin{document}

%\title{Data Normalization Using Group Orbit Optimization \thanks{This
%        work was supported by.}}

%\title{Data Normalization by Group Orbit Optimization}
\title{Multilinear Map Layer: Prediction Regularization by Structural
Constraint}
% \author{Shuchang Zhou\thanks{}
% % \thanks{Google,  Beijing, China 100084}
%         \and Zhihua Zhang\thanks{Shanghai Jiao Tong University}}

\author{
\name Shuchang Zhou     \\
\addr State Key Laboratory of Computer Architecture \\
Institute of Computing Technology \\
Chinese Academy of Sciences, Beijing, China\\
and Megvii Inc. \\
\texttt{shuchang.zhou@gmail.com} \\
          \AND
\name Yuxin Wu \\
 \addr Megvii Inc., Beijing, China, 100190 \\
              \texttt{ppwwyyxxc@gmail.com}
}

\maketitle

\section{Abstract}
In this paper we propose and study a technique to impose structural constraints
on the output of a neural network, which can reduce amount of computation and
number of parameters besides improving prediction accuracy when the output is
known to approximately conform to the low-rankness prior. The technique
proceeds by replacing the output layer of neural network with the so-called MLM layers,
which forces the output to be the result of some Multilinear Map, like a
hybrid-Kronecker-dot product or Kronecker Tensor Product. In particular, given
an ``autoencoder'' model trained on SVHN dataset, we can construct a new model
with MLM layer achieving 62\% reduction in total number of parameters and
reduction of $\ell_2$ reconstruction error from 0.088 to 0.004. Further
experiments on other autoencoder model variants trained on SVHN
datasets also demonstrate the efficacy of MLM layers.

\section{Introduction}
To the human eyes, images made up of random values are typically easily
distinguishable from those images of real world. In terms of
Bayesian statistics, a prior distribution can be constructed to describe the
likelihood of an image being a natural image. An early example of such ``image
prior'' is related to the frequency spectrum of an image, which assumes that
lower-frequency components are generally more important than high-frequency
components, to the extent that one can discard some high-frequency
components when reducing the storage size of an image, like in JPEG
standard of lossy compression of images \citep{wallace1991jpeg}.

Another family of image prior is related to the so called sparsity
pattern\citep{candes2006stable, candes2006robust, candes2008enhancing}, which
refers to the phenomena that real world data can often be constructed from a handful of exemplars modulo some negligible noise. In particular, for
data represented as vector $\bf{d} \in\FB^{m}$ that exhibits sparsity, we can
construct the following approximation:
\begin{align}
\bf{d} \approx \D \x \text{,}
\end{align}
where $\D\in\FB^{m\times n}$ is referred to as dictionary for $\bf{d}$ and $\x$ is
the weight vector combining rows of the dictionary to reconstruct $\bf{d}$.
In this formulation, sparsity is reflected by the phenomena that number of
non-zeros in $\x$ is often much less than its dimension, namely:
\begin{align}
\nnz{\x} \ll m
\text{.}
\end{align}

The sparse representation $\x$ may be derived in the framework of dictionary
learning\citep{olshausen1997sparse, mairal2009supervised} by the following
optimization:
\begin{align}
\label{align:dictionary_learning}
\min_\x \|\M - \sum_i x_i \D_i \|_F + \lambda f(\x)
\text{,}
\end{align}
where $\D_i$ is a component in dictionary, $f$ is used to induce sparsity in
$\x$, with $\ell_1$ being a possible choice.

\subsection{Low-Rankness as Sparse Structure in Matrices}

When data is represented as a matrix $\M$, the formulation of
\ref{align:dictionary_learning} is related to the rank of $\M$ by
Theorem~\ref{thm:svd-opt} in the following sense:
\begin{align}
\rank(\M) & = \min_{\x} \|\x\|_0 \\
& \text{ s.t. } \M = \sum_i x_i \D_i, \\
& \D_i = \u_i \bf{v}_i^{\top},\, \u_i^\top \u_i = 1,\, \bf{v}_i^\top \bf{v}_i =
1
\text{.}
\end{align}

Hence a low rank matrix $\M$ always has a sparse representation \wrt some rank-1
orthogonal bases. This unified view allows us to generalize the sparsity pattern
to images by requiring the underlying matrix to be low-rank.

When an image has multiple channels, it may be represented as a tensor
$\T\in\FB^{C\times H\times W}$ of order 3. Nevertheless, the matrix structure
can still be recovered by unfolding the tensor along some dimension.

\begin{figure*}
\subfigtopskip = 0pt
\begin{center}
\centering
\subfigure[The unfolding of an RGB sample image.]{\includegraphics[height=40mm]{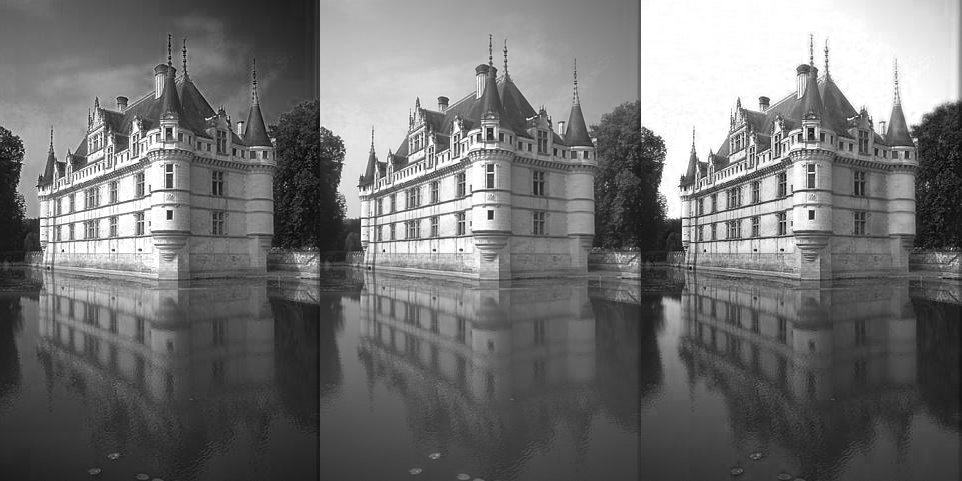}}
\subfigure[The singular values]{\includegraphics[height=40mm]{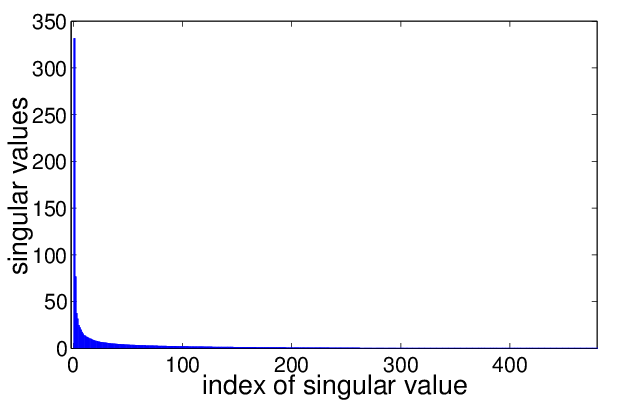}}
\end{center}
   \caption{An illustration of the singular values of the unfolding of an RBG sample image.}
\label{fig:lowrank}
\end{figure*}

In Figure \ref{fig:lowrank}, it is shown that the unfolding of the RGB image
tensor along the width dimension can be well approximated by a low-rank matrix as energy is
concentrated in the first few components.
%because of the empirical power law the singular values are conforming to.

%---------------------------------Figure---------------------------------%
%---------------------------------Figure---------------------------------%

In particular, given Singular Value Decomposition of a matrix $\M =
\U \D \V^*$, where $\U, \V$ are unitary matrices and $\D$ is a diagonal matrix with the diagonal made
up of singular values of $\M$, a rank-$k$ approximation of $\M\in\FB^{m\times
n}$ is:
\begin{align}
\M \approx \M_k = \tilde{\U} \tilde{\D} \tilde{\V}^*,\,
\tilde{\U}\in\FB^{m\times k}\text{, }\tilde{\D}\in\FB^{k\times
k}\text{, }\tilde{\V}\in\FB^{n\times k}
\text{,}
\end{align}

where $\tilde{\U}$ and $\tilde{\V}$ are the first $k$-columns of the $\U$ and
$\V$ respectively, and $\tilde{\D}$ is a diagonal matrix made up of the largest
$k$ entries of $\D$.

In this case approximation by SVD is optimal in the sense that the following
holds \citep{horn1991topics}:
\begin{align}
\M_r = \inf_{\X} \|\X - \M\|_F \text{ s.t. } \rank(\X) \le r \text{,}
\end{align}
and
\begin{align}
\M_r = \inf_{\X} \|\X - \M\|_2 \text{ s.t. } \rank(\X) \le r \text{.}
\end{align}

An important variation of the SVD-based low rank approximation is to also model
the $\ell_1$ noise in the data:
\begin{align}
\|\M\|_{RPCA} = \inf_{\S} \|\M - \S\|_{*} + \lambda \|\S\|_1
\text{.}
\end{align}

This norm which we tentatively call ``RPCA-norm'' \citep{Cand:2011}, is
well-defined as it is infimal convolution \citep{rockafellar2015convex} of two
norms.

\subsection{Low Rank Structure for Tensor}
Above, we have used the low-rank structure of the unfolded matrix of a rank-3
tensor corresponding to RGB values of an image to reflect the low rank
structure of the tensor.
However, there are multiple ways to construct of matrix $\D \in \RB^{m\times 3n}$, e.g., by stacking $\R$, $\G$ and $\B$ together horizontally.
An emergent question  is if we construct a matrix $\D' \in \RB^{3m\times n}$ by
stacking $\R$, $\G$ and $\B$ together vertically.
Moreover, it would be interesting to know if there is a method that can exploit
both forms of constructions.
It turns out that we can deal with the above variation by enumerating all
possible unfoldings. For example, the nuclear norm of a tensor may be
defined as:

\begin{definition}
\label{def:tensor_nuclear_norm}
Given a tensor $\EM$, let $\E_{(i)} = \fold^{-1}_i (\EM)$ be unfolding of $\EM$
to matrices. The nuclear norm of $\EM$ is defined w.r.t.\ some weights
$\beta_i$ (satisfying $\beta_i \ge 0$) as a weighted sum of the matrix nuclear norms of unfolded matrices $\E_{(i)}$ as:
\begin{eqnarray}
\norm{ \EM }{*} = \sum_i \beta_i \norm{ \E_{(i)} }{*} \textrm{.}
\end{eqnarray}
\end{definition}

Consequently, minimizing the tensor nuclear norm will minimize the matrix nuclear norm of all unfoldings of tensor. Further, by adjusting the weights used in the definition of the tensor nuclear norm, we may selectively minimize the nuclear norm of some unfoldings of the tensor.

\subsection{Stronger Sparsity for Tensor by Kronecker Product Low Rank}

When an image is taken as matrix, the basis for low rank factorizations are
outer products of row and column vectors. However, if data exhibits Principle of
Locality, then only adjacent rows and columns can be meaningfully related,
meaning the rank in this decomposition will not be too low. In contrast, local
patches may be used as basis for low rank structure
\citep{elad2006image, yang2008image, schaeffer2013low, dong2014compressive,
yoon2014motion}.
Further, patches may be grouped before assumed to be of low-rank
\citep{buades2005image, hu2015patch, kwok2015colorization}.

A simple method to exploit the low-rank structure of the image patches is based
on Kronecker Product SVD of matrix $\M\in\FB^{m\times n}$:
\begin{align}
\label{align:kpsvd_2}
\M =\sum_{i=1}^{\rank(\mathscr{R}(\M))} \sigma_i \U_i \otimes \V_i
\text{,}
\end{align}
where $\mathscr{R}(\A)$ is the operator
defined in \citep{van2000ubiquitous, van1993linear} which shuffles indices.
\begin{align}
\mathscr{R}(\A\otimes\B) = \vect{\A} (\vect{\B})^\top
\text{.}
\end{align}

Note that outer product is a special case of Kronecker product when
$\A\in\FB^{m\times 1}$ and $\B\in\FB^{1\times n}$, hence we have SVD as a
special case of KPSVD. The choice of shapes of $\A$ and $\B$, however,
affects  the extent to which the underlying sparsity assumption is valid.
Below we give an empirical comparison of a KPSVD with SVD. The image is of width
480 and height 320, and we approximate the image with KPSVD and SVD
respectively for different ranks. To make the results comparable, we let
$\B\in\FB^{16x20}$ to make the number of parameters in two approach equal.

%---------------------------------Figure---------------------------------%
\begin{figure}
\subfigtopskip = 0pt
\begin{center}
\centering
\subfigure[Original image selected from BSD500 dataset\citep{amfm_pami2011}]
{\includegraphics[height=24mm, width=32mm]{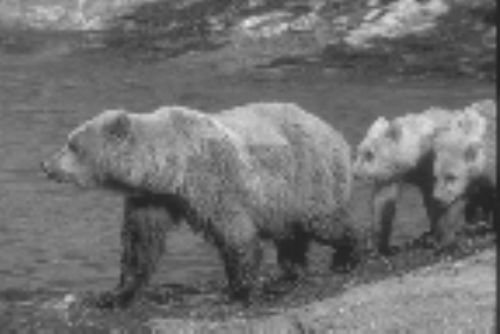}}
\subfigure[SVD approximations with rank 1, 2, 5, 10, 20 respectively from left
to right]
{\includegraphics[height=20mm, width=140mm]{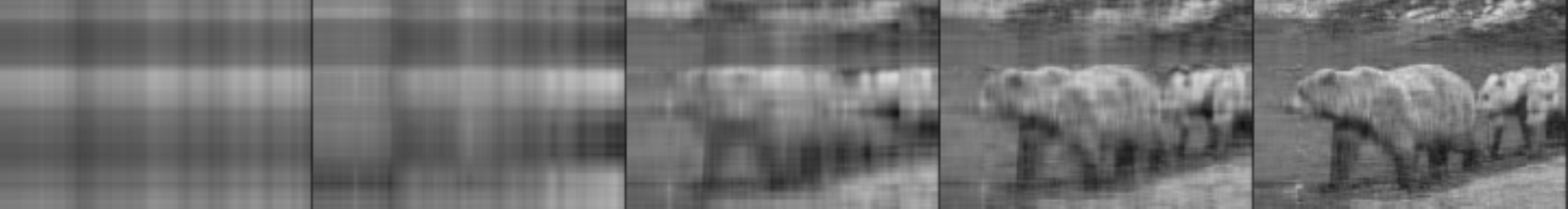}}
\subfigure[KPSVD approximations with right matrix having shape 16x20 and with
rank 1, 2, 5, 10, 20 respectively from left to right]
{\includegraphics[height=20mm, width=140mm]{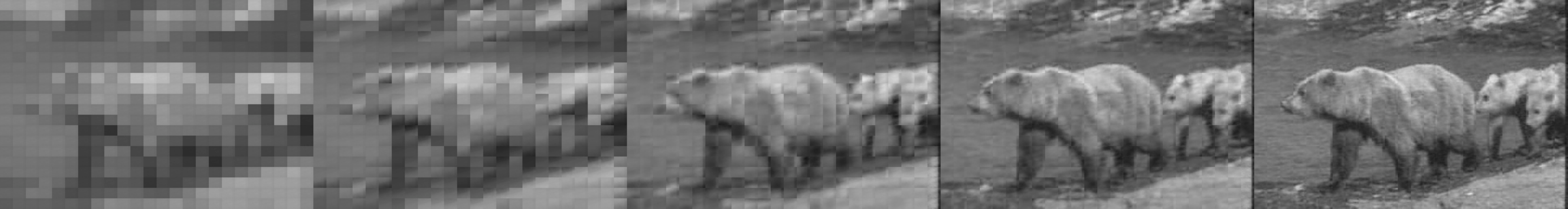}}

\end{center}
   \caption{This figures visually compares the results of KPSVD and SVD
   approximation given the same number of parameters. For this example, it
   can be seen that KPSVD with right matrix shape 16x20 is considerably better
   than SVD in approximation.}
\label{fig:svd_compare_kpsvd_bear}
\end{figure}
%---------------------------------Figure---------------------------------%

We may extend Kronecker product to tensors as Kronecker Tensor Product as:
\begin{definition}
\label{def:kp-tensor}%[Kronecker product of tensor]
Kronecker tensor product\citep{phan2013tensor, phan2012revealing} is defined for
two tensors of the same order $k$.
I.e., for two tensors:
\begin{align}
\mathcal{A}\in\FB^{m_1\times m_2,\cdots,\times m_k}
\text{,}
\end{align}
and
\begin{align}
\mathcal{B}\in\FB^{n_1\times n_2,\cdots,\times n_k}
\text{,}
\end{align}
we define Kronecker product of tensor as
\begin{align}
(\mathcal{A} \otimes \mathcal{B})_{i_1,i_2,\cdots,i_k} \coloneqq A_{\floor{i_1
/ m_1}, \floor{i_2/ m_2}, \cdots, \floor{i_k / m_k}} B_{i_1\bmod
m_1,i_2\bmod m_2,\cdots,i_k\bmod m_k}
\end{align}
where
\begin{align}
\mathcal{A} \otimes \mathcal{B}\in\FB^{m_1 n_1\times m_2 n_2,\cdots,\times
m_k n_k}
\text{.}
\end{align}

\end{definition}

With the help of Kronecker Tensor Product, \citep{phan2013tensor,
phan2012revealing} is able to extend KPSVD to tensors as:
\begin{align}
\label{align:kpsvd_tensor}
\T =\sum_{i=1}^{\rank(\mathscr{R}(\T))} \sigma_i \mathcal{U}_i \otimes
\mathcal{V}_i
\text{,}
\end{align}
where $\mathscr{R}(\T)$ is a matrix.

\section{Prediction Regularization by Structural Constraint}
In a neural network like
\begin{align}
\inf_\theta d(\Y, f(\X;\theta)) \text{,}
\end{align}

If $f(\X;\theta)$, the prediction of the network, is known to be like an image,
it is desirable to use the image priors to help improve the prediction quality. An example of such a neural network is the so-called
``Autoencoder'' \citep{vincent2008extracting, deng2010binary, vincent2010stacked,
ng2011sparse}, which when presented with a possibly corrupted image, will output
a reconstructed image that possibly have the noise suppressed.

One method to exploit the prior information is to introduce an extra
cost term to regularize the output as
\begin{align}
\label{align:label_regularized_neural_network}
\inf_\theta d(\Y, \Z) + \lambda r(\Z) \text{ s.t. } \Z = f(\X;\theta) \text{,}
\end{align}

The regularization technique is well studied in the matrix and tensor completion
literature \citep{Liu:2013:TCE:2412386.2412938,tomioka2010extension, gandy2011tensor,
signoretto2011tensor, kressner2013low, Bach:2012:OSP:2185817.2185818}. For
example, nuclear norm $\|\Z\|_*$, which is sum of singular values, or logarithm
of determinant $\log(\det(\epsilon \I + \Z \Z^\top))$ \citep{fazel2003log}, may
be used to induce low-rank structure if $\Z$ is a matrix. It is even possible to use RPCA-norm to better handle the possible sparse non-low-rank components in $\Z$ by letting
$r(\Z) = \|\Z\|_{RPCA}$.

However, using extra regularization terms also involve a few subtleties:
\begin{enumerate}
  \item The training of neural network incurs extra cost of computing the
  regularizer terms, which slows down training. This impact is exacerbated if
  the regularization terms cannot be efficiently computed in batch, like when
  using the nuclear norm or RPCA-norm regularizers together with the popular
 mini-batch based Stochastic Gradient Descent training algorithm
 \citep{bottou2010large}.
  \item The value of $\lambda$ is application-specific and may only be found
  through grid search on validation set. For example, when $r$ reflects
  low-rankness of the prediction, larger value of $\lambda$ may induce result of
  lower-rank, but may cause degradation of the reconstruction quality.
\end{enumerate}

We take an alternative method by directly restricting the parameter space of the
output. Assume in the original neural network, the output is given by a
Fully-Connected (FC) layer as:
\begin{align}
\L_{a} = h(\L_{a-1} \M_a + \b_a) \text{.}
\end{align}
It can be seen that the output $\L_{a} \in \FB^{m\times n}$ has $mn$ number of
free parameters. However, noting that the product of two matrices
$\A\in\FB^{m\times r}$ and $\B\in\FB^{r\times n}$ will have the property
\begin{align}
\rank(\A \B) \le r
\text{,}
\end{align}
when $m\ge r$ and $n\ge r$.

Hence we may enforce that $\rank(\L_a)\le r$ by
the following construct for example:
\begin{align}
\L_{a} = h(\L_{a-1} \M_a + \b_a) h(\L_{a-1} \N_a + \c_a)
\text{,}
\end{align}
when we have:
\begin{align}
\L_{a-1} \M_a + \b_a \in \FB^{m\times r} \\
\L_{a-1} \N_a + \c_a \in \FB^{r\times n}
\text{.}
\end{align}

In a Convolutional Neural Network where intermediate data are represented as
tensors, we may enforce the low-rank image prior similarly. In fact, by
proposing a new kind of output layer to explicitly encode the low-rankness of the output based on some kinds of Multilinear Map, like a
hybrid of Kronecker and dot product, or Kronecker tensor product, we are able to
increase the quality of denoising and reduce amount of computation at the same
time. We outline the two formulations below.

Assume each output instance of a neural network is an image represented by a
tensor of order 3:
\begin{align}
\T \in\FB^{C\times H\times W}
\text{,}
\end{align}
where $C$, $H$ and $W$ are number of channels, height and width of the image
respectively.

\subsection{KTP layer}
In KTP layers, we approximate $\T$ by Kronecker Tensor Product of two tensors.
\begin{align}
\label{align:kronecker-tensor-product}
\T \approx \mathcal{A} \otimes \mathcal{B}
\text{.}
\end{align}

As by applying the shuffle operator defined in \citep{van2000ubiquitous,
van1993linear}, \ref{align:kronecker-tensor-product} is equivalent to:
\begin{align}
\label{align:kronecker-tensor-product}
\mathscr{R}(\T) \approx \vect{\mathcal{A}} \vect{\mathcal{B}}^\top
\text{,}
\end{align}
hence we are effectively doing rank-1 approximation of the matrix
$\mathscr{R}(\T)$. A natural extension would then be to increase the number of
components in approximation as:
\begin{align}
\label{align:kronecker-tensor-product-ranks}
\T \approx \sum_{i=1}^K \mathcal{A}_i \otimes \mathcal{B}_i
\text{.}
\end{align}

We may further combine the multiple shape formulation of
\ref{align:kronecker-tensor-product-ranks} to get the general form of KTP layer:
\begin{align}
\label{align:kp-multiple-shape-multiple-sum}
\T \approx \sum_{j=1}^{J} \sum_{i=1}^{K} \AM_{ij} \otimes \BM_{ij}
\text{.}
\end{align}
where $\{\AM_{ij}\}_{i=1}^K$ and $\{\BM_{i,j}\}_{i=1}^K$ are of the same shape
respectively.

\subsection{HKD layer}
In HKD layers, we approximate $\T$ by the following multilinear map between
$\mathcal{A}\in\FB^{C_1\times H_1\times W_1}$ and
$\mathcal{B}\in\FB^{C_1\times C_2 \times H_2\times W_2}$, which is a hybrid of
Kronecker product and dot product:
\begin{align}
\label{align:tensor-kronecker-dot}
T_{c,h,w} \approx \tilde{T}_{c,h,w} = \tilde{T}_{c,h_1 + H_1 * h_2, w_1 + W_1 *
w_2} = \sum_{c_1} A_{c_1, h_2, w_2} B_{c,c_1,h_1,w_1}
\text{,}
\end{align}
where $h=h_1 h_2$, $w=w_1 w_2$.

The rationale behind this construction is that the Kronecker product along the
spatial dimension of $H$ and $W$ may capture the spatial regularity of the
output, which enforces low-rankness; while the dot product along $C$ would
allow combination of information from multiple channels of $\mathcal{A}$ and
$\mathcal{B}$.

In the framework of low-rank approximation, the formulation
\ref{align:tensor-kronecker-dot} is by no means unique. One could for example
improve the precision of approximation by introducing multiple components as:
\begin{align}
T_{c,h,w} = T_{c,h_1 + H_1 * h_2, w_1 + W_1 * w_2} \approx \sum_{k} \sum_{c_1} A_{k, c_1,
h_2, w_2} B_{k, c,c_1,h_1,w_1}
\text{.}
\end{align}
Hereafter we would refer to layers constructed following
\ref{align:tensor-kronecker-dot} as HKD layers. The general name of MLM layers
will refer to all possible kinds of layers that can be constructed from other
kinds of multilinear map.

\subsection{General MLM layer}
A Multilinear map is a function of several variables that is
linear separately in each variable as:
\begin{align}
f\colon V_1 \times \cdots \times V_n \to W
\text{,}
\end{align}
where $V_1,\ldots,V_n$ and $W\!$ are vector spaces with the following property:
for each $i\!$, if all of the variables but $v_i\!$ are held constant, then
$f(v_1,\ldots,v_n)$ is a linear function of $v_i\!$\citep{lang2002algebra}. It is
easy to verify that Kronecker product, convolution, dot product are all special
cases of multilinear map.

\begin{figure}
\begin{center}
  \includegraphics[width=0.6\textwidth]{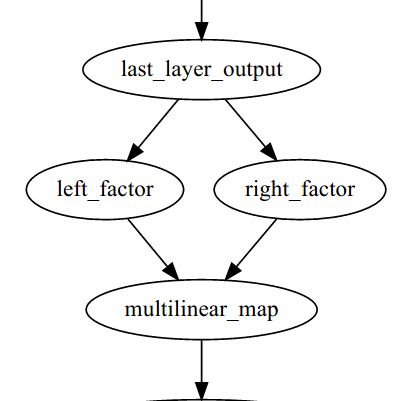}
\end{center}
   \caption{Diagram of the structure of a MLM layer. the output of the
   preceding layer is fed into two nodes where left factor tensor and right
   factor tensor are computed. Then a multilinear map is applied on the left
   and right factors to construct the output.}
\label{fig:mlm_layer}
\end{figure}

Figure~\ref{fig:mlm_layer} gives a schematic view of general structure of the
MLM layer. The left factor and right factor are produced from the same input,
which are later combined by the multilinear map to produce the output.
Depending on the type of multilinear map used, the MLM layer will become HKD
layer or KTP layer. We note that it is also possible to introduce more
factors than two into an MLM layer.

%---------------------------------Figure---------------------------------%
%---------------------------------Figure---------------------------------%

\section{Empirical Evaluation of Multilinear Map Layer}
We next empirically study the properties and efficacy of the Multilinear Map
layers and compare it with the case when no structural constraint is imposed on
the output.

\begin{figure}
\subfigtopskip = 0pt
\begin{center}
\centering
\subfigure[Original image patches selected from SVHN
dataset]{\includegraphics[height=24mm, width=130mm]{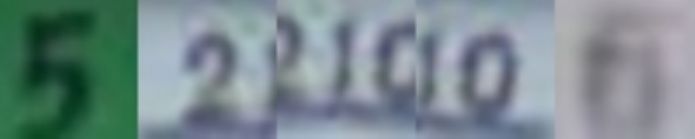}} \subfigure[Results of an Autoencoder A1]{\includegraphics[height=24mm,
width=130mm]{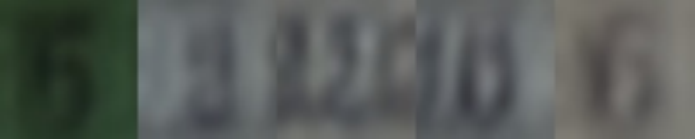}}
\subfigure[Results of an Autoencoder A2 with less neurons in the bottleneck
layer than A1]{\includegraphics[height=24mm,
width=130mm]{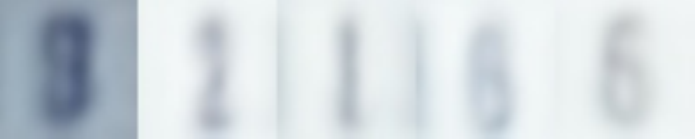}}
\subfigure[Results of an Autoencoder with as many neurons in the bottleneck
layer as A2 but uses HKD as output layer]{\includegraphics[height=24mm,
width=130mm]{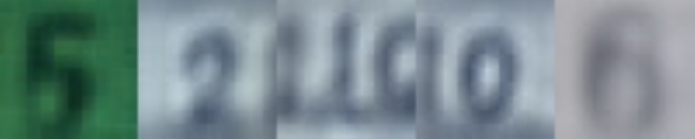}}
\end{center}
   \caption{This figures show the results of passing five cropped patched from
   SVHN dataset as input images through Autoencoders with different output
   layers.
   The first row contains the original images. The second row contains the
   output of an Autoencoder ``A1''.
   The third row contains the output of an Autoencoder ``A2", which has
   smaller number of hidden units in the bottleneck layer. The fourth row contains the
  output of an Autoencoder ``A3'' constructed from ``A2'' by replacing the output
  FC layer with a HKD layer. It can be seen that ``A3'', though with smaller
  number of hidden units in bottleneck layer, visually performs better than
  ``A1'' and ``A2'' in reconstruting the input images.}
\label{fig:svhn_autoencoder}
\end{figure}

To make a fair comparison, we first train a covolutional autoencoder
with the output layer being a fully-connected layer as a baseline. Then we
replace the fully-connected layer with different kinds of MLM layers and train
the new network until quality metrics stabilizes.
For example, Figure~\ref{fig:svhn_autoencoder} gives a subjective comparison of
HKD layer with the original model on SVHN dataset.
We then compare MLM layer method with the baseline model in terms of number of parameters and prediction
quality. We do the experiments based on implementation of MLM layers in
Theano\citep{bergstra+al:2010-scipy, Bastien-Theano-2012} framework.

%---------------------------------Figure---------------------------------%

\begin{table}[!ht]
\label{tab:eval_mlm_svhn}
  \newcommand{\twocell}[2][c]{%
  \begin{tabular}[#1]{@{}c@{}}#2\end{tabular}}

\centering \small
\caption{Evaluation of MLM layers on SVHN digit reconstruction}
  \begin{tabular}{c|c|c|c|c}
    \hline
    model & \twocell{bottleneck\\ \#hidden unit} & total \#params & layer \#params & L2 error \\\hline

    conv + FC &512 & 13.33M & 5764800 &3.4e-2\\\hline
    conv + HKD & 512 &	4.97M	&	46788&		5.2e-4\\\hline
    \twocell{conv + HKD \\multiple components} & 512 &	5.05M&		118642&	3.6e-4\\ \hline
    \hline

    conv + FC   &64&	13.40M	&	5764800&		1.3e-1\\\hline
  conv+ HKD & 64 &	5.04M	&	46788	&	1.8e-3\\\hline
  conv + KTP & 64 & 5.04M & 46618 & 3.2e-3 \\\hline
  \hline

  conv      & 16 & 5.09M  & 0 & 4.0e-2 \\\hline
  conv + FC   &16&	13.36M&		5764800	&	8.8e-2\\\hline
  conv + HKD &16& 5.00M	& 46788		& 3.8e-3 \\\hline
  conv + KTP & 16 & 5.00M & 46618 & 5.8e-3 \\\hline

\end{tabular}
\end{table}

Table~\ref{tab:eval_mlm_svhn} shows the performance of MLM layers on training an
autoencoder for SVHN digit reconstruction.
The network first transforms the $40 \times 40$ input image to a bottleneck feature vector through
a traditional ConvNet consisting of 4 convolutional layers, 3 max pooling layers and 1 fully connected layer.
Then, the feature is transformed again to reconstruct the image through 4 convolutional layers, 3 un-pooling layers and
the last fully-connected layer or its alternatives as output layer.
The un-pooling operation is implemented with the same approach used in \citep{dosovitskiy2014learning},
by simply setting the pooled value at the top-left corner of the pooled region, and leaving other as zero.
The fourth column of the table is the number of parameters in fully-connected
layer or its alternative MLM layer.
By varying the length of bottleneck feature and using different alternatives for
FC layer, we can observe that both HKD and KTP layer are good alternatives for
FC layer as output layer, and they also both significantly reduce the number of parameters.
We also tested the case with convolutional layer as output layer, and the result still shows the efficacy of MLM layer.
%---------------------------------Figure---------------------------------%

As the running time may depend on particular implementation details of the KFC
and the Theano framework, we do not report running time. However, the complexity analysis suggests
that there should be significant reduction in amount of computation.

\section{Related Work}
In this section we discuss related work not covered in previous sections.

Low rank approximation has been a standard tool in restricting the space of the
parameters. Its application in linear regression dates back to
\citep{anderson1951estimating}. In \citep{sainath2013low, liao2013large, xue2013restructuring, zhang2014extracting, denton2014exploiting}, low rank
approximation of fully-connected layer is used; and \citep{jaderberg2014speeding,
rigamonti2013learning, DBLP:journals/corr/LebedevGROL14, lebedev2014speeding,
denil2013predicting} also considered low rank approximation of convolution
layer. \citep{zhang2014efficient} considered approximation of multiple layers
with nonlinear activations. To our best knowledge, these methods only consider
applying low-rank approximation to weights of the neural network, but not to
the output of the neural network.

As structure is a general term, there are also other types of structure that
exist in the desired prediction. Neural network with structured prediction also
exist for tasks other than autoencoder, like edge detection\citep{dollar2013structured}, image
segmentation\citep{zheng2015conditional, farabet2013learning}, super
resolution\citep{dong2014image}, image generation\citep{dosovitskiy2014learning}.
The structure in these problem may also exhibit low-rank structure exploitable by MLM layers.

\section{Conclusion and Future Work}
In this paper, we propose and study methods for incorporating the low-rank image
prior to the predictions of the neural network. Instead of using
regularization terms in the objective function of the neural network, we
directly encode the low-rank constraints as structural constraints by requiring the output of
the network to be the result of some kinds of multilinear map. We consider a few
variants of multilinear map, including a hybrid-Kronecker-dot product and Kronecker tensor product.
We have found that using the MLM layer can
significantly reduce the number of parameters and amount of computation for
autoencoders on SVHN.

As future work, we note that when using
$\ell_1$ norm as objective together with the structural constraint, we could
effectively use the norm defined in Robust Principal Value Analysis as our
objective, which would be able to handle the sparse noise that may otherwise
degrade the low-rankness property of the predictions.
In addition, it would be interesting to investigate applying the structural
constraints outlined in this paper to the output of intermediate layers of
neural networks.

\bibliographystyle{spmpsci}
\bibliography{thesis}

\end{document}